\documentclass[10pt,twocolumn,letterpaper]{article}

\usepackage{cvpr}              
\usepackage[accsupp]{axessibility}  
\usepackage{graphicx}
\usepackage{amsmath}
\usepackage{amssymb}
\usepackage{booktabs}
\usepackage{multirow}
\usepackage{booktabs}
\usepackage{pifont}
\usepackage[T1]{fontenc}
\renewcommand{\thefootnote}{\fnsymbol{footnote}}
\makeatletter
\newcommand{\setword}[2]{%
  \phantomsection
  #1\def\@currentlabel{\unexpanded{#1}}\label{#2}%
}
\makeatother

\newcommand*{\affaddr}[1]{#1} 
\newcommand*{\affmark}[1][*]{\textsuperscript{#1}}
\newcommand*{\email}[1]{\texttt{#1}}
\newcommand{\cmark}{\ding{51}}
\newcommand{\xmark}{\ding{55}}

\usepackage[dvipsnames]{xcolor}

\newcommand*{\Scale}[2][4]{\scalebox{#1}{$#2$}}%
\usepackage[pagebackref,breaklinks,colorlinks]{hyperref}

\usepackage[capitalize]{cleveref}
\crefname{section}{Sec.}{Secs.}
\Crefname{section}{Section}{Sections}
\Crefname{table}{Table}{Tables}
\crefname{table}{Tab.}{Tabs.}

\def\cvprPaperID{2465} 
\def\confName{CVPR}
\def\confYear{2022}

\begin{document}

\title{vCLIMB: A Novel Video Class Incremental Learning Benchmark}

\author{%
Andrés Villa\affmark[1,2]$^\ast$, Kumail Alhamoud\affmark[2], Victor Escorcia\affmark[3], Fabian Caba Heilbron\affmark[4], \\Juan León  Alcázar\affmark[2], Bernard Ghanem\affmark[2]
\\
\normalsize \small\affaddr{\affmark[1]Pontificia Universidad Católica de Chile}, \small\affaddr{\affmark[2]King Abdullah University of Science and Technology (KAUST)}, \\ \normalsize \small\affaddr{\affmark[3]Samsung AI Center Cambridge}, \small\affaddr{\affmark[4]Adobe Research}
\\
\small \email{afvilla@uc.cl, kumail.hamoud@kaust.edu.sa, v.castillo@samsung.com}\\
\small \email{caba@adobe.com, \{juancarlo.alcazar, bernard.ghanem\}@kaust.edu.sa}
}
\maketitle

\renewcommand\thefootnote{$\ast$}
\footnotetext{Work done during an internship at KAUST.}

\begin{abstract}
Continual learning (CL) is under-explored in the video domain. The few existing works contain splits with imbalanced class distributions over the tasks, or study the problem in unsuitable datasets. We introduce vCLIMB, a novel video continual learning benchmark. vCLIMB is a standardized test-bed to analyze catastrophic forgetting of deep models in video continual learning. In contrast to previous work, we focus on class incremental continual learning with models trained on a sequence of disjoint tasks, and distribute the number of classes uniformly across the tasks.
We perform in-depth evaluations of existing CL methods in vCLIMB, and observe two unique challenges in video data. The selection of instances to store in episodic memory is performed at the frame level. Second, untrimmed training data influences the effectiveness of frame sampling strategies. We address these two challenges by proposing a temporal consistency regularization that can be applied on top of memory-based continual learning methods. Our approach significantly improves the baseline, by up to 24\% on the untrimmed continual learning task. The code of our benchmark can be found at: \href{https://vclimb.netlify.app/}{https://vclimb.netlify.app/}.


\end{abstract}

\section{Introduction}
\label{sec:intro}

Deep neural networks rely on large-scale datasets to achieve state-of-the-art performance on modern computer vision tasks\cite{deng2009imagenet, krizhevsky2012imagenet, lin2014microsoft, carreira2017quo}. A significant result of such pre-training is enabling feature reuse \cite{yosinski2014transferable, donahue2014decaf}, by means of fine-tuning the learned weights for smaller downstream tasks \cite{long2015fully, ren2015faster, bertinetto2016fully}. Due to legal or technical constraints, and the fact that labeling data is expensive and time-consuming \cite{su2012crowdsourcing}, real-world deep-learning pipelines would rarely involve a single fine-tuning stage. Instead, these pipelines could require the sequential fine-tuning of large models in a set of independent tasks that are learned sequentially.
Under these conditions, deep neural networks suffer from what is known as \textit{catastrophic forgetting} \cite{french1999catastrophic}, where the fine-tuning on novel tasks significantly reduces the performance of the model in a previously learned task, and \textit{drift} \cite{royer2015classifier}, where unseen training data does not fit the previously estimated class distribution. Continual learning \cite{dhar2019learning} directly models such a scenario, by adapting a neural network model into a sequential series of tasks. We focus on a special case of CL: \textit{class incremental learning (CIL)}, where the labels and data are mutually exclusive between tasks, training data is available only for the current task, and there are no tasks ids.

With 500 hours of video from diverse categories uploaded to YouTube every minute and a billion people actively using TikTok every month \cite{chityoutube, citetiktok}, video content of varying quality is available at an unprecedented scale. With such large volumes of data, it is important to develop models that can effectively learn from continuous streams of untrimmed video data. 
Remarkably, few research efforts have addressed continual learning with video \cite{park2021class, ma2021class, zhao2021video}. Despite these current works, video continual learning methods still show a large variability in their experimental protocols, making direct comparisons hard to establish. These protocols present the following limitations. \textbf{(1)} They are not publicly available. \textbf{(2)} They do not explore a realistic setup with untrimmed videos. \textbf{(3)} Most of them leverage a large pretraining step, which warms up the model by learning a sample of classes from the same distribution and is not always available in a real continual learning scenario.

We directly address these limitations and propose vCLIMB (\textbf{v}ideo \textbf{CL}ass \textbf{I}ncre\textbf{M}ental Learning \textbf{B}enchmark), a novel benchmark devised for the evaluation of continual learning in video. Our test-bed defines a fixed task split on the original training and validation sets of three well known video datasets: UCF101\cite{soomro2012ucf101}, ActivityNet\cite{caba2015activitynet} and Kinetics \cite{carreira2017quo}. vCLIMB follows the standard class incremental continual learning scenario, but includes some modifications to better fit the nature of video data in human action recognition tasks. First, to achieve fair comparisons between video CIL methods that use memory, we re-define \textit{memory size} \cite{rebuffi2017icarl, Wu_2019_CVPR} to be the total number of frames, instead of the total number of video instances, that the memory can store. We report this as \textit{Memory Frame Capacity} in our tables to avoid confusion with the memory size defined in image CIL. This means we are not only concerned about selecting the best videos to store in memory, but  we also want to identify the best set of frames to keep in memory. Second, since fine-grained temporal video annotations are expensive (especially for long videos), we analyze the effect of using trimmed and untrimmed video data in continual learning. To the best of our knowledge, this is the first work to explore continual learning with untrimmed videos.


Using vCLIMB's data splits, we establish an initial set of baselines by adapting well-known continual learning methods\cite{rebuffi2017icarl, Aljundi_2018_ECCV, kirkpatrick2017overcoming, Wu_2019_CVPR} from the image recognition domain into the activity recognition domain. After benchmarking these baseline methods, we propose a novel strategy for video continual learning that leverages the inherent temporal consistency in video data to better approach the continual action recognition problem. We believe that vCLIMB's standardized approach to prototyping and evaluating video continual learning will enable future works in this area.




\vspace{2pt}\noindent\textbf{Contributions.} This paper proposes vCLIMB, a novel benchmark for continual learning in video, which focuses on the activity classification task. Our work brings the following contributions: \textbf{(1)} A standardized benchmark for continual learning in video action recognition, which defines the training protocols and associated metrics for three video datasets; this novel continual learning setup includes a more realistic combination of trimmed and untrimmed videos. \textbf{(2)} We re-purpose and evaluate four baseline methods from the image domain into the video domain. \textbf{(3)} We present a novel strategy based on consistency regularization that can be built on top of memory-based methods to reduce memory consumption while improving performance.


\section{Related Work}
\label{sec:sota}
Continual Learning (CL) studies methods that can learn novel concepts continually without forgetting previous knowledge. In the class incremental learning (CIL) setting, models are trained sequentially on a set of tasks. Each incremental task consists of labeled data from novel classes, which are learned individually without considering the data from previous tasks. CIL approaches can be grouped into two broad categories: regularization-based and memory-based methods. While regularization-based methods penalize abrupt changes in the most relevant parameters learned from previous tasks, memory-based methods mitigate catastrophic forgetting by retaining a limited amount of training instances from previous tasks. Although there is extensive literature on continual learning \cite{shin2017continual, kirkpatrick2017overcoming, rebuffi2017icarl, kemker2017fearnet, zenke2017continual, Aljundi_2018_ECCV, chaudhry2018riemannian, ostapenko2019learning, cermelli2020modeling, liu2020mnemonics}, we restrict our review to a sample of the most widely adopted baselines.

\begin{table*}[t]
    \centering
    \small
    \resizebox{17cm}{!}{
    \begin{tabular}{l l r r r r r r c}
        \toprule
         \multicolumn{1}{c}{\multirow{2}{*}{\bf Set}}&
         \multicolumn{1}{c}{\multirow{1}{*}{\bf In-distribution }}&
         \multicolumn{1}{c}{\multirow{2}{*}{\bf Tasks}}
         & \multicolumn{3}{c}{\bf Videos Per Task} & \multicolumn{1}{c}{\multirow{2}{*}{ \begin{tabular}[c]{@{}c@{}}\textbf{Classes} \\ \textbf{Per Task} \end{tabular}}} & \multicolumn{1}{c}{\multirow{2}{*}{ \begin{tabular}[c]{@{}c@{}}\textbf{Avg. Frames} \\ \textbf{Per Video} \end{tabular}}} & \multicolumn{1}{c}{\multirow{2}{*}{ \begin{tabular}[c]{@{}c@{}}\textbf{Untrimmed} \\ \textbf{Video} \end{tabular}}} \\
            & \bf Pretraining & & Train & Val & Test &  &  &   \\
        \toprule
        i-Something-Something-B0 \cite{zhao2021video} & None & 4 & -- & -- & -- & 10 & -- & \xmark \\
        i-Something-Something-B20 \cite{zhao2021video} & 20 classes & 5 & -- & -- & -- & 4 & -- & \xmark\\
         i-Kinetics-B0 \cite{zhao2021video} & None & 4 & -- & -- & -- & 10 & -- & \xmark\\
        i-Kinetics-B20 \cite{zhao2021video} & 20 classes & 5 & -- & -- & -- & 4 & -- & \xmark\\
        UCF101-50 \cite{park2021class}  & 51 classes & 5/10/25 & -- & -- & -- & 10/5/2 & -- & \xmark\\
        HMDB51-25 \cite{park2021class}  & 26 classes & 5/25 & -- & -- & -- & 5/1 & -- & \xmark\\
        Something-Something V2-90 \cite{park2021class} & 84 classes & 9/18 & -- & -- & -- & 10/5 & -- & \xmark\\
        \toprule
        vCLIMB UCF101 & None & 10 & 928 & 131 & 272  & 10 & 183 & \xmark\\
        vCLIMB UCF101 & None & 20 & 464 & 65 & 136  & 5 & 183 & \xmark\\
        vCLIMB Kinetics & None & 10 & 24628 & 1988 & 3977 & 40 & 250 & \xmark\\
        vCLIMB Kinetics & None & 20 & 12314 & 994 & 1988 & 20 & 250 & \xmark\\
        vCLIMB ActivityNet-Untrim. & None & 10 & 1001 & 492 & -- & 20 & 3542 &  \cmark \\
        vCLIMB ActivityNet-Untrim. & None & 20 & 500 & 246 & -- & 10 & 3542 &  \cmark \\
        vCLIMB ActivityNet-Trim. & None & 10 & 1541 & 765 & -- & 20 & 3879 &  \xmark \\
        vCLIMB ActivityNet-Trim. & None & 20 & 770 & 383 & -- & 10 & 3879 &  \xmark \\
        \bottomrule
    \end{tabular}
    }
    \caption{
    \textbf{CIL Benchmark Statistics.} In vCLIMB we provide 8 splits for class incremental learning, each split contains 10 or 20 tasks. Our Kinetics and ActivityNet setups contain large-scale video data and provide long sequences of tasks with no video pretraining. This makes our splits more suitable for measuring forgetting. We highlight that ActivityNet-Untrim provides a realistic challenge to test a model’s ability to continually learn from weakly labeled video streams. 
    }
    \vspace{-4mm}
    \label{tab:1}
\end{table*}

\vspace{2pt}\noindent\textbf{Regularization-Based Methods.} Regularization techniques try to keep constant the weights that are important for the previous tasks. They differ in how they estimate the importance value of the model parameters learned in previous tasks. These values are commonly updated and stored in an importance matrix at the end of each task. Elastic Weight Consolidation (EWC) \cite{kirkpatrick2017overcoming} uses the Fisher Information Matrix to make that estimation. Memory Aware Synapses (MAS) \cite{Aljundi_2018_ECCV} estimates the importance of each parameter in a self-supervised manner by measuring how small changes in the parameters affect the output of the model. MAS is of particular interest in the video domain because it can handle weakly labeled instances. These methods include a regularization factor $\lambda_{reg}$, which controls how relevant the previously learned tasks are. For a large factor, the model will prioritize the previous tasks over the current one.



\vspace{2pt}\noindent\textbf{Memory-Based Methods.} Memory-based methods select and store samples into a memory buffer for future replay. This memory has a limited size and is available when learning a new task. While a naive baseline randomly chooses instances from previously learned classes, current strategies attempt to select the best subset of training samples per class to move into the buffer. We focus on two representative methods that have been shown to work well with high-resolution image datasets. Incremental Classifier and Representation Learning (iCaRL) \cite{rebuffi2017icarl} combines rehearsal and distillation strategies. During training, it selects the most representative instances per class following a nearest-mean-of-exemplars rule. The Bias Correction method (BiC) \cite{Wu_2019_CVPR} follows iCaRL's instance sampling approach, but augments the classification layer with a bias correction layer. Bias correction mitigates the imbalance between the large amount of data from the new task and the relatively scarce data from previous tasks, which is only available in memory. 

\vspace{2pt}\noindent\textbf{Consistency Regularization.} Consistency regularization techniques are used to ensure that a model’s output is invariant to various augmentations, which is common in semi-supervised learning\cite{sajjadi2016regularization, laine2017temporal, xie2020unsupervised, berhelot2019mixmatch, sohn2020fixmatch}. For example, such methods have been shown to improve the generation of images from a few samples\cite{zhang2020consistency, zhao2020diffaugment}. Our work investigates catastrophic forgetting in video continual learning and proposes a consistency loss to help memory-replay methods significantly alleviate this impeding nuisance.



\vspace{2pt}\noindent\textbf{Continual Learning in Video.} 
Despite the growing interest in CL in the image domain, the first three works to report results on video data were only recently published. Zhao \etal\cite{zhao2021video} proposed a spatio-temporal knowledge transfer strategy to mitigate catastrophic forgetting. A concurrent work \cite{park2021class} estimated the subset of feature channels that contributes the most to the predictions of the previous tasks, and introduced a temporal mask that keeps this subset stable while learning a new task. It also included a distillation loss, which allows only the least relevant feature maps to be updated while learning a new task. Finally, Ma \etal\cite{ma2021class} also approached the class incremental video classification problem by regularizing the feature space in consecutive tasks.

Although existing works bootstrapped the study of continual learning in video data, all of them use different evaluation protocols, making direct comparisons between methods difficult. Moreover, these works propose to pretrain the model with a large set of classes of the same data distribution (up to half of the total), as shown in Table \ref{tab:1}. Such an arrangement is unnatural for continual learning, as it makes it difficult to disentangle the effects of catastrophic forgetting. In contrast, our benchmark presents a more challenging and realistic setup. Our splits contain up to 20 tasks, with a balanced number of classes per task. Furthermore, vCLIMB includes three video datasets, which enables the study of the continual learning problem in more diverse scenarios. Finally, previous works do not provide a detailed analysis of the proposed video continual learning setups. In our benchmark, we provide extensive empirical evaluations to analyze individual splits and identify unique properties of continual learning in video, including the memory size and the adoption of untrimmed video configurations.





\section{vCLIMB: A Video Class Incremental Learning Benchmark} 

\begin{figure*}[t]
    \vspace{-1.2cm}
    \centering
    \includegraphics[width=0.9\linewidth]{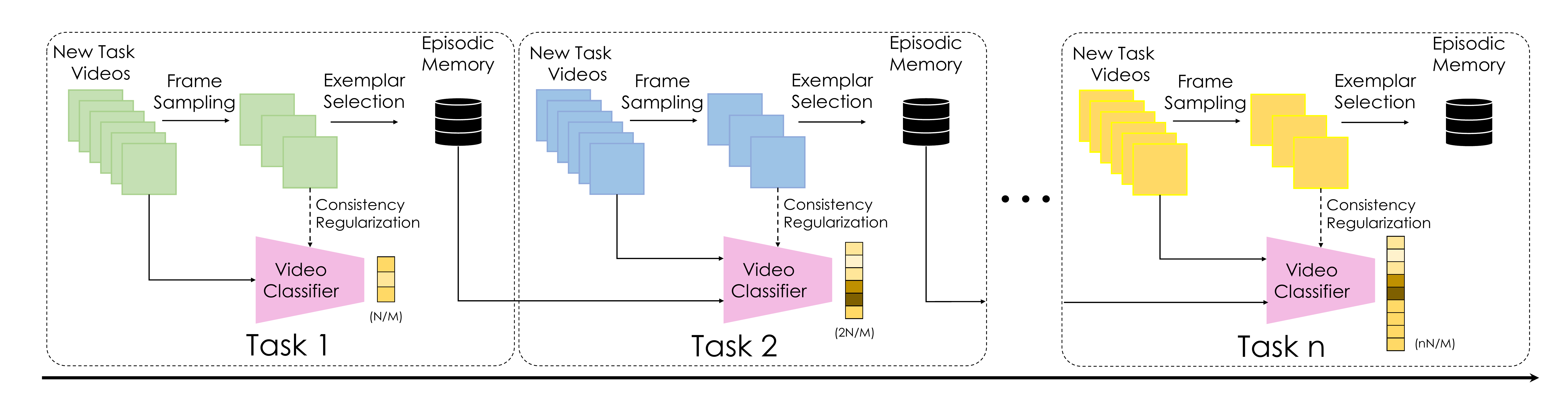}
    \caption{
    \textbf{Outline of our Memory-based CIL setting.} In video CIL, a model sequentially learns a set of video classification tasks. Memory-based methods define a limited episodic memory to store a few temporally down-sampled examples of previous tasks. We propose a consistency regularization that helps the model to better remember previous tasks from down-sampled examples in memory.
    }
    \vspace{-4mm}
    \label{fig:teaser_fig}
\end{figure*}



\paragraph{Notation \& Problem Definition} Similar to the image domain, we train a single neural network ($F_{\omega}$) with parameter set ($\omega$) over a sequence of tasks ($R_{n}$).  Each task in the sequence contains its own training data $R_{i} = \{(X_{0}, Y_{0}), (X_{1}, Y_{1})... (X_{n}, Y_{n})\}$, with $X_{n}$ the input data and $Y_{n}$ its corresponding ground truth. We optimize the parameter set $\omega$ in $F$ sequentially over $R_{n}$, searching for $\bar F_{\omega}$ that maximizes the average accuracy over $R_{n}$.

\vspace{-0.2cm}
\paragraph{Datasets and Tasks.} The bottom half of Table~\ref{tab:1} summarizes the main attributes of vCLIMB.  We design  vCLIMB on top of three well-known datasets for action recognition: \textbf{(1)} UCF101\cite{soomro2012ucf101}, which has 13.3K videos from  101 classes, \textbf{(2)} Kinetics\cite{carreira2017quo}, a large-scale video dataset with over 300K short clips distributed over 400 action classes, and \textbf{(3)} ActivityNet\cite{caba2015activitynet}, which can be used for trimmed and untrimmed activity classification and contains 20K videos from  200 activity classes. We utilize the diversity of videos in ActivityNet and provide two subsets: ActivityNet-Trim, in which every part of the videos when an action happens is considered as an independent video, and ActivityNet-Untrim, which is more challenging and labels a whole video with its most representative action class. We create two different CIL sequences of tasks for each dataset. The first sequence contains ten tasks, and the second sequence contains twenty tasks. Statistics about the number of classes per task, the number of videos per task, and the average number of frames in each split are provided in Table~\ref{tab:1}.

\vspace{3pt}\noindent\textbf{Metrics for Video Continual Learning (CL).} In vCLIMB, we use the standard CL metrics: Final Average Accuracy (Acc) and Backward Forgetting (BWF). Acc is the average classification accuracy of the model evaluated on all learned tasks, including the last task it was trained on \cite{NIPS2017_f8752278,wolczyk2021continual}. This metric is essential to show how the average performance of the model degrades as it learns new tasks. BWF complements Acc and measures the influence of the learned task $i$ in the performance of the previous tasks \cite{NIPS2017_f8752278}, as: 
\begin{equation}
BWF_i = \frac{1}{N_i-1} \sum_{j=1}^{N_i-1} (R_{j,j} - R_{N_i,j})
\end{equation}
where $N_i$ is the number of learned tasks after learning the task $i$, and $R_{j,j}$ and $R_{N_i,j}$ represents the accuracy on the task $j$ after learning the task $j$, and learning a new task the task $i$, respectively. Given a total of $N$ tasks, we report the final Backward Forgetting $BWF=BWF_N$ in our tables. 


\subsection{Unique Challenges of Video CL}
\label{subsec:challenges}
Video CIL comes with unique challenges. (1) Memory-based methods developed in the image domain are not scalable to store full-resolution videos, so novel methods are needed to select representative frames to store in memory. (2) Untrimmed videos have background frames that contain less helpful information, thus making the selection process more challenging. (3) The temporal information is unique to video data, and both memory-based and regularization-based methods need to mitigate forgetting while also integrating key information from this temporal dimension. These challenges informed the design choices of vCLIMB.

\vspace{-10pt}

\paragraph{Re-defining Memory Size.} Unlike image benchmarks for class incremental learning (CIL), our video instances contain a temporal dimension whose size could show large variability. To favor fair comparisons between methods, we define the working memory size in terms of stored frames. This design choice avoids a scenario where longer videos are preferred (or even trivially selected) to maximize the amount of data stored in the working memory. Moreover, it creates a new unique aspect of CIL in video data, as methods must decide first what subset of frames should be selected, and then decide what video to store according to the sub-sampled videos. We follow the same instance per class ratio as \cite{rebuffi2017icarl}, which is 20 when the model has learned all training classes. Therefore, we define a memory that can save at most 8000, 4000, and 2020 videos for Kinetics, ActivityNet, and UCF101, respectively. If we store videos in memory without down-sampling them, this would correspond to saving 3.25\%, 25.95\%, and 21.76\% of the total frames in Kinetics, ActivityNet, and UCF101. As we show later, these storage requirements can be significantly reduced using temporal consistency regularization. 



\vspace{-0.5cm}
\paragraph{Untrimmed Video Data for CIL.} In the untrimmed classification setting, the action of interest may occur at any time in the video and its boundaries are not known. This problem formulation has no direct counterpart in the image classification domain. The annotation scheme of ActivityNet allows us to analyze this scenario for class incremental learning. In ActivityNet, videos contain one or multiple temporal segments defining the occurrence of an action instance, while unlabeled segments constitute a background set where no relevant action takes place \cite{caba2015activitynet}.

We leverage this unique property of video data and define two independent setups for video CIL. In the trimmed setup, we only use frames that belong to a labeled action segment. In the untrimmed setup, we freely sample frames from the whole video, regardless of whether they belong to the main action or the background. For consistency in the untrimmed scenario, we give every frame in the video the same label. We select a primary label as the action with the longest temporal support in the video, and discard any video that contains instances of 2 or more different labels. We empirically find that this assignment only discards 0.15\% of the ActivityNet dataset. This untrimmed learning task more closely resembles the real world scenario of CIL, where a model learns from a continuous stream of diverse videos. Given the scale of current video services and the costly nature of fine-grained labels, real models would likely be learning from a stream with weak video-level annotations. 

\vspace{-0.5cm}
\paragraph{Baselines}
We implement and evaluate these four continual learning methods to serve as baselines \cite{kirkpatrick2017overcoming, rebuffi2017icarl, Wu_2019_CVPR, Aljundi_2018_ECCV} because they are widely used and easily scalable to the video domain. We also compare with a naive memory-based strategy, which selects samples randomly for memory creation. We provide the implementations for these methods as part of our video CIL benchmark. 

\subsection{A Stronger Baseline with Temporal-consistency Regularization}
\label{sec:temp_consistency}
We present a novel strategy for CIL in the video domain. Our approach relies on one of the unique characteristics of video data, temporal resolution consistency. Figure \ref{fig:teaser_fig} illustrates our pipeline for memory-based video CIL methods. After completing the training on videos from the first task, the model has the option to select a few temporally sub-sampled examples to store in the episodic memory. When learning the second task, the model is trained on both the new task videos and a few past-task examples it retained in memory. We introduce a regularization loss, represented by the dashed arrow in Figure \ref{fig:teaser_fig}, during the fine-tuning phase of every new task. This loss constrains the network to estimate similar feature representations for the original video clip and a temporally down-sampled version of it. 

\begin{figure}[t]

\centering
\includegraphics[width=0.9\linewidth]{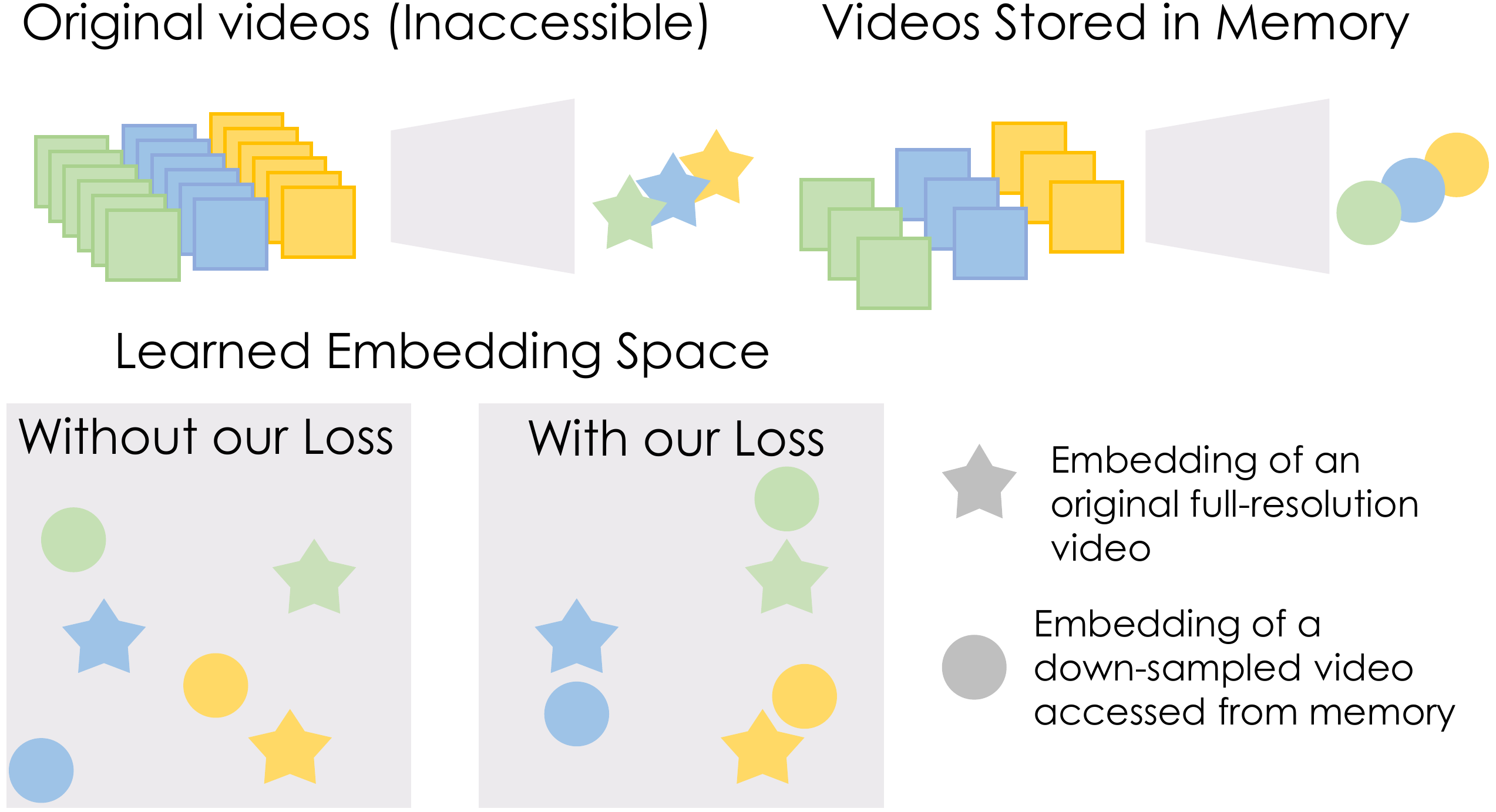}
\vspace{-0mm}
\caption{
\textbf{Temporal Consistency Loss.} Our consistency loss encourages learning representations that are robust to frame sampling, enabling the CIL model to remember old tasks by looking only at down-sampled versions of those videos.
}
\vspace{-4mm}
\label{fig:cons_fig}
\end{figure}

This constraint aims to optimize the effectiveness of the data stored in the episodic memory, by enforcing a similar representation between the original clip and its temporally down-sampled version. As shown in Figure \ref{fig:cons_fig}, the most relevant aspects of our loss is that it reduces the drift between the samples stored in memory and the original samples used at training time. While simple, our constraint directly addresses one of the key aspects of CIL in video: its episodic memory is composed of frame sets instead of full-resolution video clips. In practice, we enforce similar representations by adding a regularization term to the loss function. 

\textbf{Temporal Consistency Loss.} When training on a new task, each video ($X$) will have an augmented version ($X^{d}$) by means of temporal down-sampling. We use pairs of $X$ and $X^d$ to calculate a consistency loss $L_{c}$ over a single forward pass of the network $F$:
\begin{equation}
    L_{c} = (1 - \lambda) L_{cls} (F(X), Y) + \lambda L_{cls} (F(X^{d}), Y),
\end{equation}
where $L_{cls}$ is the cross-entropy loss, $Y$ is the ground truth label of $X$, and $\lambda$ is the consistency regularization factor. 

Our consistency regularization strategy is model agnostic and thus can be adapted to any backbone and CIL memory-replay strategy. Since we use the same set of weights ($F$) for both loss terms, our method only introduces a linear time increase in the total number of FLOPs.

\section{Experimental Evaluation}
\label{sec:cliv}
We now proceed with the experimental assessment of the class incremental continual learning task in vCLIMB. In this section, we first discuss the details of the re-implementation of image methods for continual learning in the video domain. Then, we proceed with the empirical assessment of these baseline methods on the three datasets included in the vCLIMB benchmark: UFC101\cite{soomro2012ucf101}, Kinetics\cite{carreira2017quo} and ActivityNet \cite{caba2015activitynet}. For the ActivityNet dataset, we also evaluate the CIL task in the presence of trimmed and untrimmed video annotations. Moreover, we assess the effectiveness of the proposed regularization method in all the previous scenarios and datasets.




\vspace{-0.15cm}
\paragraph{Implementation Details.} The memory-based \cite{rebuffi2017icarl, Wu_2019_CVPR} and regularization-based \cite{Aljundi_2018_ECCV, kirkpatrick2017overcoming} baselines are trained for 50 and 20 epochs, respectively, following a fully-supervised setup on each CIL task. We use TSN \cite{TSN} with a ResNet-34 backbone pretrained on ImageNet. We follow the same temporal data augmentation proposed in \cite{TSN}, using $N=8$ segments per video. We optimize our model using Adam \cite{Adam14} with a learning rate of $1\times10^{-3}$. For the temporal consistency loss factor, we use $\lambda=0.5$. We run MAS \cite{Aljundi_2018_ECCV} with a regularization factor $\lambda_{reg}$ of $3\times10^5$. We found EWC \cite{kirkpatrick2017overcoming} to be more challenging to tune for video CIL. The best results are obtained by picking a different regularization factor $\lambda_{reg}$ for EWC on each individual dataset: $3\times10^3$ for UCF101, $5\times10^2$ for Kinetics, and $3\times10^5$ for ActivityNet.


\begin{table*}[t!]
\centering
 \small
\resizebox{17cm}{!}{
\begin{tabular}{cccccccccccccc} 
    \toprule

    \multicolumn{1}{c}{\multirow{2}{*}{\bf Model}} & 
    \multicolumn{1}{c}{\multirow{2}{*}{ \begin{tabular}[c]{@{}c@{}}\textbf{Num.} \\ \textbf{Task} \end{tabular}}}
    & \multicolumn{4}{c}{\bf Kinetics} & \multicolumn{4}{c}{\bf ActivityNet-Trim} & \multicolumn{4}{c}{\bf UCF101}
    \\\cmidrule(lr){3-14}
         & & \multicolumn{1}{c}{\multirow{2}{*}{ \begin{tabular}[c]{@{}c@{}}\textbf{Mem. Video} \\ \textbf{Instances} \end{tabular}}} & \multicolumn{1}{c}{\multirow{2}{*}{ \begin{tabular}[c]{@{}c@{}}\textbf{Mem. Frame} \\ \textbf{Capacity} \end{tabular}}} & \multicolumn{1}{c}{\bf Acc $\uparrow$ } & \multicolumn{1}{c}{\bf BWF $\downarrow$ } &
         \multicolumn{1}{c}{\multirow{2}{*}{ \begin{tabular}[c]{@{}c@{}}\textbf{Mem. Video} \\ \textbf{Instances} \end{tabular}}} &
         \multicolumn{1}{c}{\multirow{2}{*}{ \begin{tabular}[c]{@{}c@{}}\textbf{Mem. Frame} \\ \textbf{Capacity} \end{tabular}}} &
         \multicolumn{1}{c}{\bf Acc $\uparrow$ } & \multicolumn{1}{c}{\bf BWF $\downarrow$} &
         \multicolumn{1}{c}{\multirow{2}{*}{ \begin{tabular}[c]{@{}c@{}}\textbf{Mem. Frame} \\ \textbf{Capacity} \end{tabular}}} &
         \multicolumn{1}{c}{\multirow{2}{*}{ \begin{tabular}[c]{@{}c@{}}\textbf{Mem. Frame} \\ \textbf{Capacity} \end{tabular}}} & \multicolumn{1}{c}{\bf Acc $\uparrow$ } & \multicolumn{1}{c}{\bf BWF $\downarrow$} \\\\
    \midrule
    EWC & 10 & None & None & 5.81\% & 16.05\% & None & None & 4.02\% & 5.32\% & None & None & 9.51\% & 98.94\%
    \\
    MAS & 10 & None & None & 7.81\% & 10.12\% & None & None & 8.11\% & 0.18\% & None & None & 10.89\% & 11.11\%
    \\
    EWC & 20 & None & None & 2.95\% & 32.70\% & None & None & 1.28\% & 3.77\% & None & None & 4.71\% & 92.12\%
    \\
    MAS & 20 & None & None & 4.25\% & 5.54\% & None & None & 4.61\% & 0.1\% & None & None & 5.90\% & 5.31\%
    \\
    \cmidrule{1-14}
    Naive & 10 & 8000 & \Scale[0.94]{2\times10^6} & 30.14\% & 41.30\% & 4000 & \Scale[0.94]{15.5\times10^6} & 47.20\% & 20.64\% & 2020 & \Scale[0.94]{3.69\times10^5} & 91.42\% & 7.43\%
    \\
    iCaRL & 10 & 8000 & \Scale[0.94]{2\times10^6} & 32.04\% & 38.74\% & 4000 & \Scale[0.94]{15.5\times10^6} & 48.53\% & 19.72\% & 2020 & \Scale[0.94]{3.69\times10^5} & 80.97\% & 18.11\%
    \\
    BiC & 10 & 8000 & \Scale[0.94]{2\times10^6} & 27.90\% & 51.96\% & 4000 & \Scale[0.94]{15.5\times10^6} & 51.96\% & 24.27\% & 2020 & \Scale[0.94]{3.69\times10^5} & 78.16\% & 18.49\%
    \\
    Naive & 20 & 8000 & \Scale[0.94]{2\times10^6} & 23.47\% & 48.05\% & 4000 & \Scale[0.94]{15.5\times10^6} & 40.78\% & 23.18\% & 2020 & \Scale[0.94]{3.69\times10^5} & 87.40\% & 10.96\%
    \\
    iCaRL & 20 & 8000 & \Scale[0.94]{2\times10^6} & 26.73\% & 42.25\% & 4000 & \Scale[0.94]{15.5\times10^6} & 43.33\% & 21.57\% & 2020 & \Scale[0.94]{3.69\times10^5} & 76.59\% & 21.83\%
    \\
    BiC & 20 & 8000 & \Scale[0.94]{2\times10^6} & 23.06\% & 58.97\% & 4000 & \Scale[0.94]{15.5\times10^6} & 46.53\% & 15.95\% & 2020 & \Scale[0.94]{3.69\times10^5} & 70.69\% & 24.90\%
    \\
    \bottomrule
    \end{tabular}
}
\caption{
\textbf{Baseline Video CIL Results}. We report the average accuracy (Acc) and the backward forgetting (BWF) at 10 and 20 tasks from three action recognition benchmarks. Regularization-based methods (at the top of the table), under-perform memory-based methods, shown at the bottom. Consistent with results in the image domain, the longer 20-task sequences are more challenging for all methods across all dataset. We highlight that the reported improvements in the image domain do not directly translate to the video domain, as no single method obtains the best performance in every setup. 
}
\vspace{-4mm}
\label{tab:datasets_results}
\end{table*}

\subsection{Baselines for Video CIL}
As outlined in \ref{subsec:challenges} the datasets in vCLIMB differ in scale, so we set a different working memory limit for each dataset according to their total number of frames. For each dataset, we perform experiments on two different splits: a 10-task split and a 20-task split. In Table \ref{tab:datasets_results}, we report the results obtained using the best hyper-parameters for each method. 

Consistent with image class incremental learning \cite{clbenchmark}, the regularization-based methods EWC \cite{kirkpatrick2017overcoming} and MAS \cite{Aljundi_2018_ECCV} lag significantly behind replay-based methods, regardless of the difficulty of the dataset or the number of tasks. This is because regularization methods only penalize changes to the model parameters, and thus experience an unavoidable trade-off between learning new tasks and forgetting older tasks. If a larger regularization parameter is used to emphasize learning new tasks, forgetting increases and the average accuracy on old tasks is compromised. If a smaller regularization parameter is used to emphasize remembering old tasks, the accuracy on newer tasks is impaired. Given the difficulty of video CIL, this limitation highlights the importance of future exploration of more sophisticated strategies of memory-free approaches. 

Surprisingly, no memory-based approach is superior across all video datasets. While the naive baseline, which randomly samples instances for memory, outperforms the other baselines on UCF101, iCaRL and BiC are better on Kinetics and ActivityNet. We believe that fixing this discrepancy by finding memory-based methods designed specifically for video is an important research direction.

\paragraph{Number of Tasks and Forgetting.} We observe that the longer (20-task) sequence is a more challenging setup, where the average accuracy is always lower for any method on any of the three datasets. Following this trend, all memory-based methods forget more as the number of tasks in the sequence increases. Similar to the image domain, evaluating on long sequences of tasks accentuates the shortcomings of CIL methods, and is thus suitable for the study of strategies that mitigate forgetting.  We pose closing the forgetting gap between the 10-task and 20-task scenarios as an important research direction. 


\begin{table*}[t]
\centering
\small
\resizebox{17cm}{!}{
\begin{tabular}{cccccccccccc} 
    \toprule
    \multicolumn{1}{c}{\multirow{2}{*}{\bf Model}} & 
    \multicolumn{1}{c}{\multirow{2}{*}{ \begin{tabular}[c]{@{}c@{}}\textbf{Frames} \\ \textbf{per video} \end{tabular}}}
    & \multicolumn{3}{c}{\bf Kinetics} & \multicolumn{3}{c}{\bf ActivityNet-Trim} & \multicolumn{3}{c}{\bf UCF101}
    \\\cmidrule(lr){3-11}
         & & \multicolumn{1}{c}{\bf Mem. Frame Capacity} & \multicolumn{1}{c}{\bf Acc $\uparrow$ } & \multicolumn{1}{c}{\bf BWF $\downarrow$} &
         \multicolumn{1}{c}{\bf Frame Capacity} &\multicolumn{1}{c}{\bf Acc $\uparrow$} & \multicolumn{1}{c}{\bf BWF $\downarrow$} &
         \multicolumn{1}{c}{\bf Mem. Frame Capacity} & \multicolumn{1}{c}{\bf Acc $\uparrow$} & \multicolumn{1}{c}{\bf BWF $\downarrow$} \\
    \midrule
    iCaRL & 4 & \Scale[0.94]{3.2\times10^4} & 30.73\% & 40.36\% & \Scale[0.94]{1.6\times10^4} & 21.63\% & 36.98\% & \Scale[0.94]{8.08\times10^3} & 80.32\% & 17.13\% 
    \\
    iCaRL & 8 & \Scale[0.94]{6.4\times10^4} & 32.04\% & 38.48\% & \Scale[0.94]{3.2\times10^4} & 21.54\% & 33.41\% & \Scale[0.94]{16.16\times10^3} & 81.12\% & 18.25\%
    \\
    iCaRL & 16 & \Scale[0.94]{12.8\times10^4} & 31.36\% & 38.74\% & \Scale[0.94]{6.4\times10^4} & 25.27\% & 29.71\% & \Scale[0.94]{32.32\times10^3} & 81.06\% & 18.23\%
    \\
    iCaRL & ALL & \Scale[0.94]{2\times10^6} & 32.04\% & 38.74\% & \Scale[0.94]{15.5\times10^6} & 48.53\% & 19.72\% & \Scale[0.94]{3.69\times10^5} & 80.97\% & 18.11\%
    \\
    \cmidrule{1-11}
    iCaRL+TC & 4 & \Scale[0.94]{3.2\times10^4} & 35.32\% & 34.07\% & \Scale[0.94]{1.6\times10^4} & 42.99\% & 23.82\% & \Scale[0.94]{8.08\times10^3} & 73.85\% & 26.35\%
    \\
    iCaRL+TC & 8 & \Scale[0.94]{6.4\times10^4} & 36.24\% & 33.83\% & \Scale[0.94]{3.2\times10^4} & 45.73\% & 18.90\% & \Scale[0.94]{16.16\times10^3} & 74.25\% & 25.27\%
    \\
    iCaRL+TC & 16 & \Scale[0.94]{12.8\times10^4} & 36.54\% & 33.53\% & \Scale[0.94]{6.4\times10^4} & 44.04\% & 22.82\% & \Scale[0.94]{32.32\times10^3} & 75.84\% & 23.23\%
    \\
    \bottomrule
    \end{tabular}
}
\caption{
\textbf{Ablation study results with different memory sizes}. We compare iCaRL \cite{rebuffi2017icarl} with and without Temporal Consistency (iCaRL+TC) on the 10-task trimmed action recognition setups. iCaRL experiences a loss of accuracy as the memory size decreases in the challenging Kinetics and ActivityNet-Trim setups. Applying TC allows us to reduce the memory size by 2 orders of magnitude while retaining the performance in ActivityNet-Trim, and it even outperforms the version of iCaRL that uses all frames in Kinetics.
}
\vspace{-2mm}
\label{tab:frames_results}
\end{table*}

\paragraph{Relevance of Memory Size in Video.}
\label{par:relev}
UCF101 stands out in Table \ref{tab:datasets_results}, since memory-based methods perform surprisingly well in this datatset. In fact, the 91.42\% accuracy obtained by the naive rehearsal baseline on the UCF101 10-task CIL is almost on par with the 94.9\% accuracy reported for training TSN \cite{TSN}, which is the the backbone in our experiments, on all UCF101 classes simultaneously. This creates a sharp contrast with Kinetics and ActivityNet results, where the best CIL baseline achieve 32.04\% and 51.96\% respectively. In comparison, TSN trained on all the action classes of the whole dataset at once achieves 73.9\% on Kinetics \cite{xie2018rethinking} and 88\% on ActivityNet \cite{TSN}.

Park \etal \cite{park2021class} conducted experiments on UCF101 and made the observation that storing a subset of frames in memory results in a similar performance to storing the whole video \cite{park2021class}. Our hypothesis is that such good performance and apparent invariance to frame sampling is an artifact specific to the UCF101 dataset, and that it is not the norm in video class incremental learning. Our experiments on Kinetics and ActivityNet in Table \ref{tab:frames_results} show that this is indeed not the case for more challenging video datasets. In particular for ActivityNet, storing four frames per video in memory, results in an overall decrease of accuracy by 27\% compared to storing all frames. We hypothesize that the temporal dependencies in Kinetics and ActivityNet are more complex and more relevant for the task, thus models with naive sampling strategies struggle to remember older tasks from severely down-sampled videos. 


\subsection{Remembering from Down-sampled Videos.} To avoid incurring large memory requirements, it is favorable for video CIL models to down-sample videos before storing them for future replay. This means the model will learn a new task with full-resolution videos, but its memory will be composed of temporally sub-sampled versions of videos belonging to older tasks. Unfortunately, the distribution shift from the original training data to the modified stored exemplars causes a decline in the accuracy, which is evident from the Kinetics and ActivityNet results in Table~\ref{tab:frames_results}.

To mitigate forgetting, continual video action recognition models must learn robust action embeddings, which are invariant to the temporal resolution of the video. We use our temporal consistency loss to train the CIL model jointly on both full-resolution videos and temporally down-sampled videos. Our proposed strategy helps achieve this desirable invariance. Table~\ref{tab:frames_results} reports the results of using our consistency loss, which is explained in Section \ref{sec:temp_consistency}, to mitigate forgetting in models that remember from down-sampled videos. Since no single memory-based method consistently outperforms the other methods across all datasets, we choose the more established baseline iCaRL \cite{rebuffi2017icarl} to perform the sub-sampled memory experiments.

\paragraph{Consistency Regularization on Kinetics.} The last three rows of Table \ref{tab:frames_results} summarize the effect of applying our temporal regularization on limited size memories. Temporal consistency regularization (labeled TC in the table) reduces forgetting on the 10-task Kinetics split regardless of how many frames per video are stored. In particular, our best baseline (iCaRL) tested on a memory of 16, 8, or 4 frames per video is significantly improved when our temporal consistency term is added to iCaRL’s loss objective. For example, the accuracy is improved by more than 4.5\% in the scenario where 4 frames per video are stored. It is worth highlighting that adding temporal consistency enables us to store as few as 4 frames per video and yet achieve more than 3\% accuracy improvement over storing full videos, which requires about 100 times greater \textit{memory frame capacity.}

\paragraph{Consistency Regularization on Trimmed ActivityNet.} We observe a similar trend on ActivityNet with temporal consistency resulting in even more sizable improvements. Specifically, adding the regularization term with a model that has access to a memory consisting of 8 frames per video results in a massive 24\% improvement. Moreover, with 4 frames per video stored in memory, our temporally consistent model comes close to achieving similar performance to the model that uses full-resolution videos. In fact, this significantly closes the 27\% accuracy gap between storing full-resolution videos in memory and storing 4 frames per video when our regularization is not used. Our results show that our method is most relevant for datasets that require more sophisticated temporal reasoning like ActivityNet.

\begin{table*}[t]
\centering
 \small
 
\begin{tabular}{ccccccc} 
    \toprule
    \multicolumn{1}{c}{\multirow{2}{*}{\bf Model}} & \multicolumn{1}{c}{\multirow{2}{*}{ \begin{tabular}[c]{@{}c@{}}\textbf{Frames} \\ \textbf{per video} \end{tabular}}} & 
    \multicolumn{1}{c}{\multirow{2}{*}{ \begin{tabular}[c]{@{}c@{}}\textbf{Mem.} \\ \textbf{Frame Capacity} \end{tabular}}}
    & \multicolumn{2}{c}{\bf ActivityNet-Untrim} & 
    \multicolumn{2}{c}{\bf ActivityNet-Trim}
    \\\cmidrule(lr){4-7}
         & & & \multicolumn{1}{c}{\bf Acc $\uparrow$} & \multicolumn{1}{c}{\bf BWF $\downarrow$} & \multicolumn{1}{c}{\bf Acc $\uparrow$} & \multicolumn{1}{c}{\bf BWF $\downarrow$}\\
    \midrule
    iCaRL & 4 & \Scale[0.94]{1.6\times10^4} & 16.28\% & 32.75\% & 21.63\% & 36.98\%
    \\
    iCaRL & 8 & \Scale[0.94]{3.2\times10^4} & 16.67\% & 31.96\% & 21.54\% & 33.41\%
    \\
    iCaRL & 16 & \Scale[0.94]{6.4\times10^4} & 21.27\% & 28.94\% & 25.27\% & 29.71\%
    \\
    \midrule
    iCaRL+TC & 4 & \Scale[0.94]{1.6\times10^4} & 36.07\% & 22.39\% & 42.99\% & 23.82\%
    \\
    iCaRL+TC & 8 & \Scale[0.94]{3.2\times10^4} & 40.29\% & 20.80\% & \textbf{45.73\%} & 18.90\%
    \\
    iCaRL+TC & 16 & \Scale[0.94]{6.4\times10^4} & \textbf{40.45\%} & 21.21\% & 44.04\% & 22.82\%
    \\
    \bottomrule
    \end{tabular}
\caption{
\textbf{Ablation study results with trimmed and untrimmed videos}. All the experiments involve sequentially training on 10 tasks. ActivityNet-Untrim provides a more realistic and challenging setup to evaluate CIL models. We impose the strict resource constraint of 4, 8, and 16 frames per video stored in memory. Our temporal consistency approach significantly improves the accuracy of the baseline method trained on both ActivityNet setups with limited memory. The best performing setting for each data split is highlighted in the table. 
}
\vspace{-4mm}
\label{tab:untrimmed_results}
\end{table*}

\paragraph{How Many Frames Should be Stored in Memory?} 
A major challenge for adapting continual learning methods to video is that videos consume significantly more memory than images due to the added temporal dimension. Our experiments in Table \ref{tab:frames_results} show that a consistency-based training framework makes it possible to down-sample videos before storing them for future replay, resulting in remarkably small memories and minor performance degradation. In particular on the challenging dataset Kinetics, iCaRL without TC trained with full-resolution memory achieves an average accuracy of 32.04\%. Yet, adding the TC loss and training on memory of only 8 frames per video exemplar results in an even better average performance of 36.24\%. This is impressive, since the average number of frames per video in Kinetics is 250, which translates to only 3.2\% of the average Kinetics video being stored in memory. 

Our experiments on ActivityNet-Trim also show a similar trend towards alleviating large memory requirements. Since 8 frames represent 0.21\% of the average ActivityNet video, naively storing 8-frame exemplars with no temporal consistency in training results in a 27\% drop in accuracy. Using consistency regularization and 8 frames per memory sample, we are able to reduce the difference in accuracy from 27\% to only 3\%.

\vspace{-2mm}
\paragraph{Frame Sampling in UCF101.} We revisit our assertion that remembering old tasks in UCF101 is unaffected by the number of frames stored in memory. We vary the number of frames per video used: from storing all frames to storing 16, 8, and 4 frames. The results reported in Table \ref{tab:frames_results} clearly show that the performance of iCaRL is almost the same in all these different scenarios, validating the claim that UCF101 is not a prototypical dataset to evaluate CIL methods. For completeness, we also run the same set of experiments using the proposed consistency regularization scheme. We notice that it does not help increase the performance. This is unsurprising for two reasons. \textbf{(i)} Class incremental learning on UCF101 is already not that much more challenging than training on the whole dataset in one task, as we showed in Section \ref{par:relev}. \textbf{(ii)} As the first half of Table \ref{tab:frames_results} shows, class incremental learning with a memory buffer on UCF101 is invariant to the number of frames per video stored in memory. This can be explained by the strong scene bias exhibited in UCF101 \cite{choi2019sdn}. Thus, incorporating the TC loss in this dataset, which has a very small accuracy gap between CIL training and fully-supervised training, is not expected to improve accuracy.


\paragraph{UCF101 for video CIL.} In conclusion, we still recommend using the UCF101 splits for prototyping video CIL methods. However, due to its simplicity, we encourage the community to evaluate new video CIL methods on the more challenging splits from Kinetics and ActivityNet.

\begin{figure}[!htp]

\centering
\includegraphics[width=0.9\linewidth]{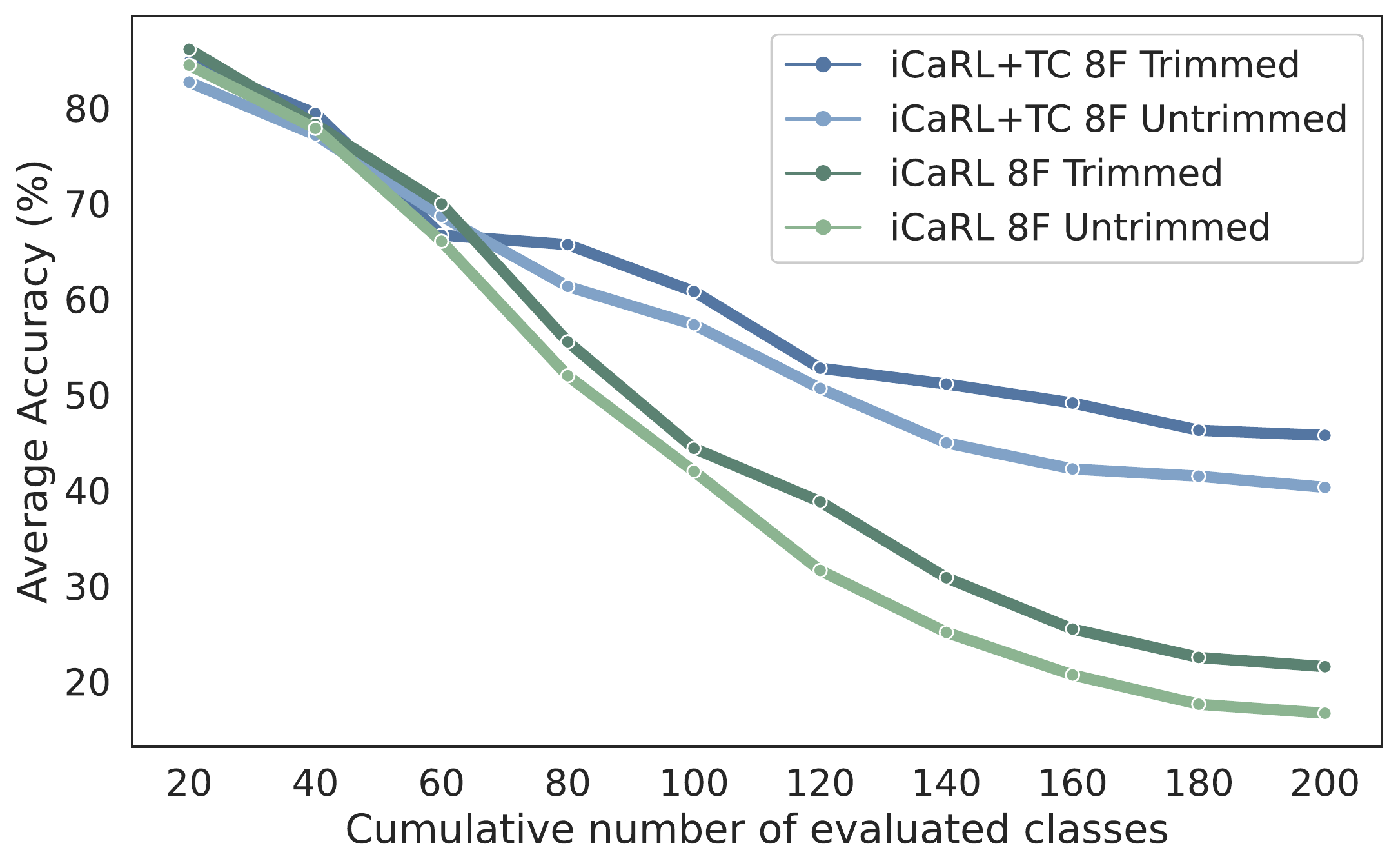}
\vspace{-3mm}
\caption{
\textbf{Average Accuracy in the Validation Set.} We train iCaRL sequentially on 10 tasks and enforce a memory constraint of 8 frames per video. Our temporal consistency (TC) loss significantly improves iCaRL's performance on ActivityNet-Trim and ActivityNet-Untrim
}
\vspace{-5mm}
\label{fig:res_actNet}
\end{figure}

\paragraph{Class Incremental Learning from Untrimmed Videos.} As observed in Table~\ref{tab:untrimmed_results}, we perform a set of experiments to evaluate the realistic class incremental learning scenario with untrimmed videos and make a few interesting observations. First, ActivityNet-Untrim is more challenging than ActivityNet-Trim. iCaRL baseline \cite{rebuffi2017icarl} achieves a better performance on ActivityNet-Trim regardless of the number of frames per video stored in memory. Second, our temporal consistency regularization improves \cite{rebuffi2017icarl} by large margins in both ActivityNet setups. The progressive performance of iCaRL on the trimmed and untrimmed setups in the 10-task case, where 8 frames are used to represent a memory video, are plotted in Figure \ref{fig:res_actNet}. Our regularization loss enhances the performance by 24\% in both cases.  

\paragraph{Why Does Consistency Regularization Work?}
We hypothesize that the large accuracy gains garnered by adding the consistency regularization loss are due to two reasons. First, regularization may make the training more stable. When training the model on augmented examples, we expect the model to learn more robust representations that can make learning new tasks easier. Second, consistency regularization forces the backbone to learn action embeddings that are invariant to the number of frames used to represent the video. Current video CIL methods store down-sampled videos in the episodic memory. However, without applying the consistency loss, the model naturally learns completely different features for densely sampled videos and sparsely sampled videos. Thus, the model struggles to remember the old action representations from this memory of down-sampled videos. This distribution shift between videos of different temporal resolution is especially apparent in datasets with more complicated temporal dependencies like ActivityNet. This points to why our approach improves the performance by large margins on this dataset and suggests TC loss does not skew the models towards spurious scene features, rather it manages to retain meaningful temporal features.

\section{Conclusion and Limitations}
In this paper, we propose and analyze vCLIMB, a continual learning benchmark for video action recognition. We expose and tackle the unstudied accuracy drop, which is experienced by memory-based video CIL models and is caused by frame sub-sampling. In our experiments, we sample frames uniformly and leverage a consistency loss to significantly alleviate that accuracy drop, by up to 24\% in CIL untrimmed video classification. We think that exploring non-uniform sampling strategies is another promising direction, but we leave that exploration for future work. 

\paragraph{Acknowledgments.} This publication is based upon work supported by the King Abdullah University of Science and Technology (KAUST) Office of Sponsored Research (OSR) under Award No. OSR-CRG2021-4648. Authors also thank Centro Nacional de Inteligencia Artificial CENIA,
FB210017, BASAL, ANID.

{\small
\bibliographystyle{ieee_fullname}
\bibliography{egbib}
}

\end{document}


\section{Supplementary}
We complement the empirical evaluations of section 4, and present additional in-depth analysis of the results. First we assess the Average Accuracy and the BWF individually along the full set of tasks. We conduct this analysis for Kinetics, ActivityNet-Trim and ActivityNet-Untrim. These results highlight the effectiveness of the Temporal Consistency loss as the class incremental learning tasks progress. 

\subsection{The Naive Memory-based Baseline}
Our naive memory-based baseline method is a simpler version of iCaRL. It follows iCaRL's inference approach, which is based on metric learning. The difference between the two methods, is the way they select the videos to store for future replay. Our naive baseline selects the videos randomly, while iCaRL estimates how close each video sample is to its class prototype. Thus, iCaRL is expected to perform better than the naive baseline. Experimentally, we find that the naive baseline outperforms iCaRL on UCF101. However, iCaRL achieves better results on the more challenging splits drawn from Kinetics and ActivityNet.

\subsection{Temporal Progression of Temporal Consistency (TC) Loss}
\begin{figure}[t]
     \centering
     \begin{subfigure}[b]{0.46\textwidth}
         \centering
         \includegraphics[width=\textwidth]{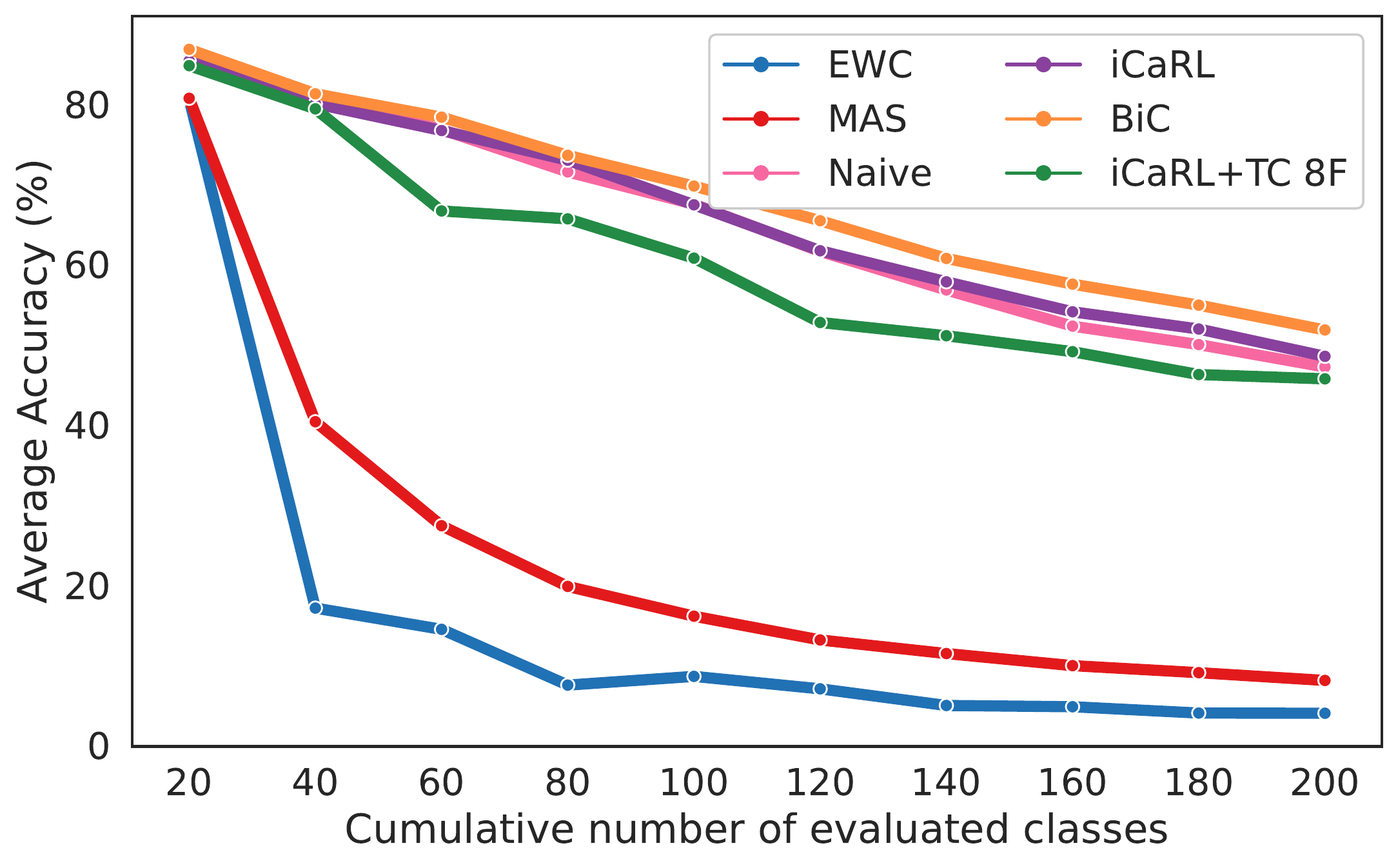}
         \caption{}
         \label{fig:acc_Model_actNet}
     \end{subfigure}
     \hfill
     \begin{subfigure}[b]{0.46\textwidth}
         \centering
         \includegraphics[width=\textwidth]{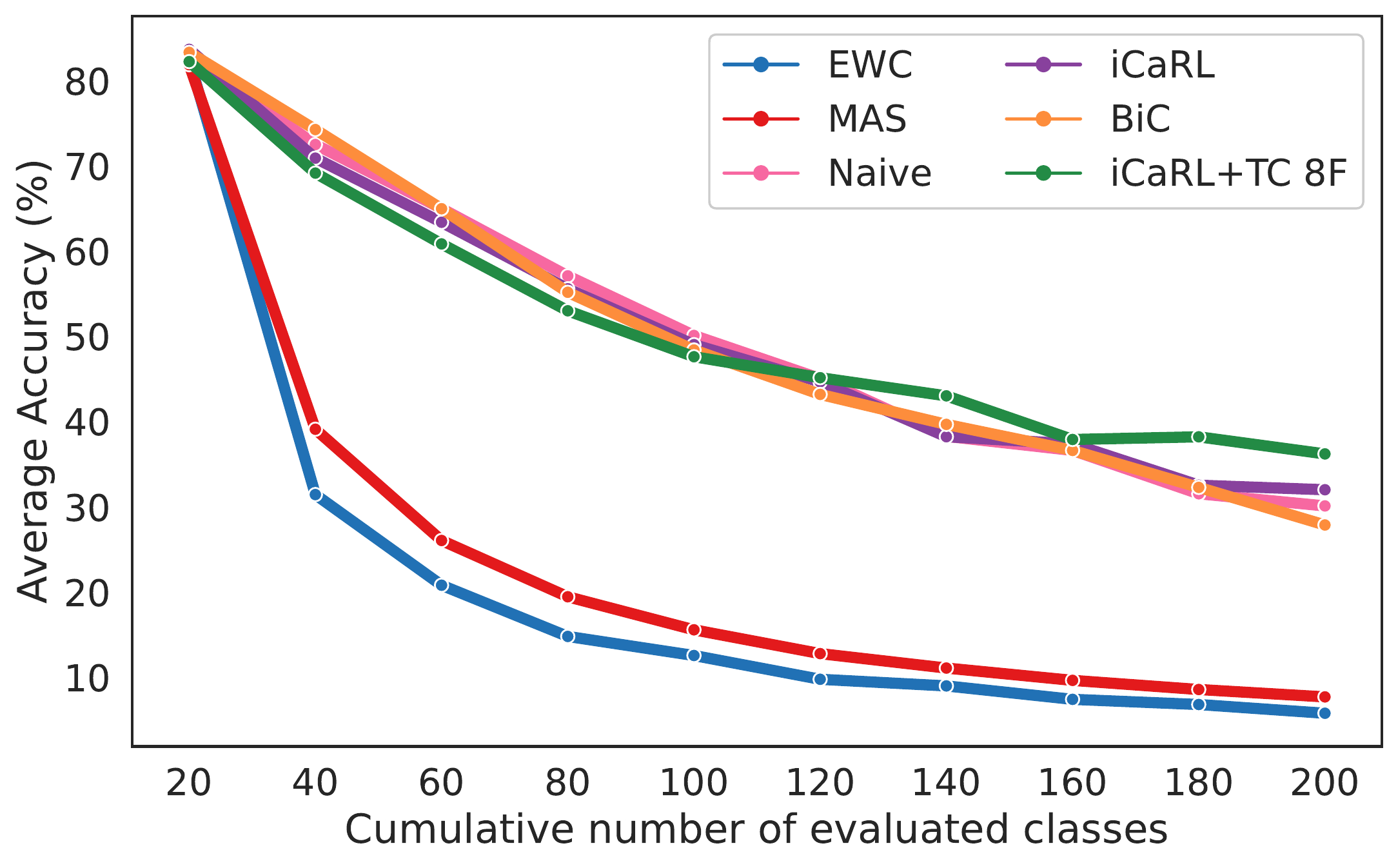}
         \caption{}
         \label{fig:acc_Model_Kinetics}
     \end{subfigure}

\caption{
\textbf{Average Accuracy in the Validation and Test Sets of ActivityNet-Trim and Kinetics.} We sequentially train different memory-based methods (iCaRL, BiC, iCaRL+TC) and regularization-based methods (EWC, MAS) on 10 tasks. Our temporal consistency (TC) loss achieves competitive results on: (a). ActivityNet-Trim and (b). Kinetics while using significantly smaller memory. Specifically, it stores around 100 times fewer frames than the other memory-based methods evaluated on Kinetics but still outperforms them.  
}
\label{fig:acc_Models}
\end{figure}

Figure \ref{fig:acc_Models} shows the Average Accuracy of the baseline methods as they sequentially learn more classes in vCLIMB. Not only does the use of Temporal Consistency (TC) reduce the required memory size considerably, but it also keeps the performance almost at the same level of methods that use full memory. In particular for ActivityNet-Trim,  \ref{fig:acc_Model_actNet} shows that most of the performance is retained using only 8 frames per video in memory when using TC. Such an improvement is notable for a dataset like  ActivityNet-Trim, which is composed of long video sequences.  Notably for kinetics \ref{fig:acc_Model_Kinetics}, TC allows for storing only 8 frames while even improving the performance. 

Moreover, our TC loss achieves outstanding results on ActivityNet-Untrim and ActivityNet-Trim with the Naive memory-based method, as shown in Table \ref{tab:naive_results}. It improves the performance by 21\% on both benchmarks when 4 and 8 frames per video are saved. This shows the capacity of our TC loss to be used with different memory-based methods.

\subsection{Sensitivity to the number of stored frames}
\begin{figure}[t]
     \centering
     \begin{subfigure}[b]{0.46\textwidth}
         \centering
         \includegraphics[width=\textwidth]{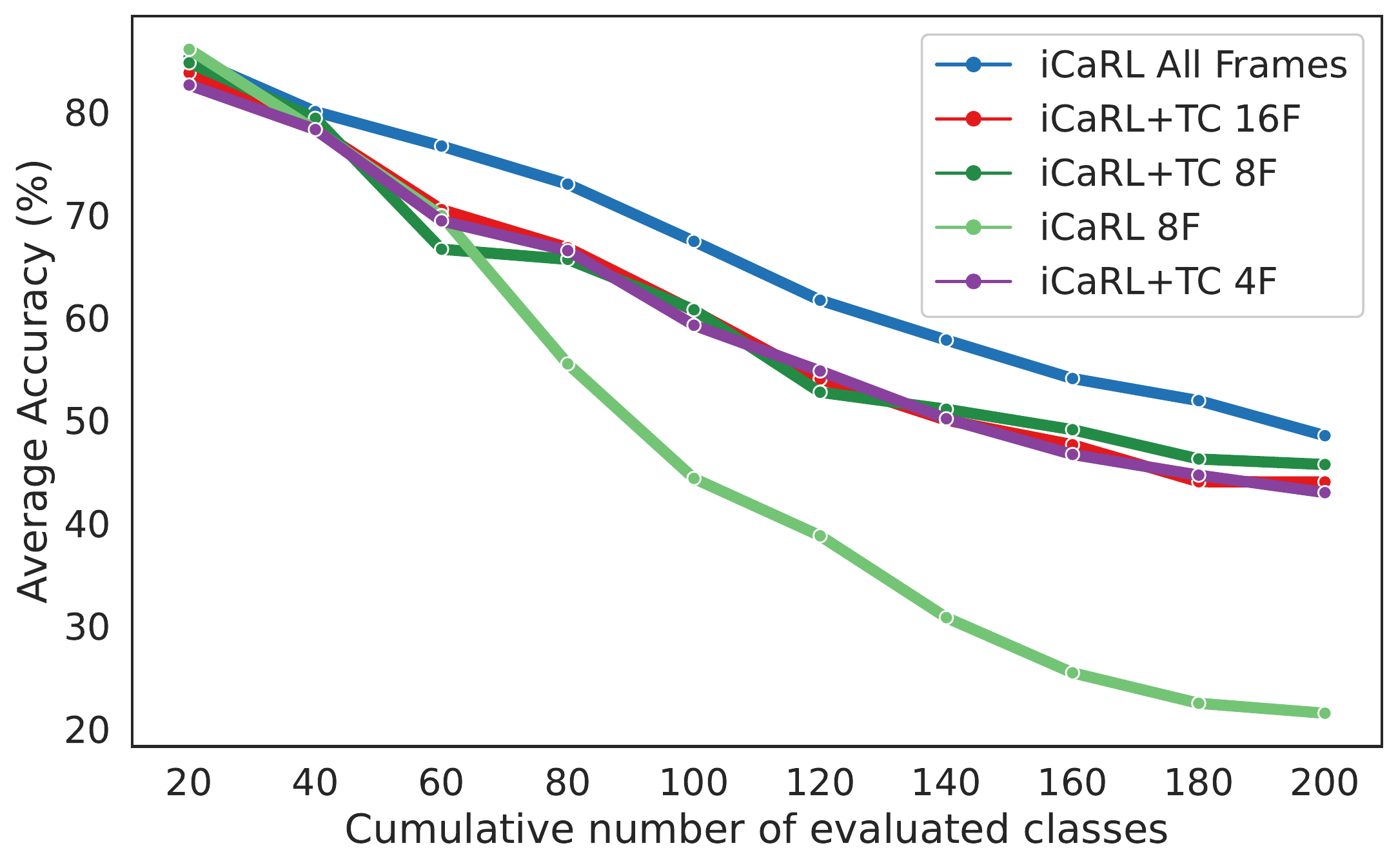}
         \caption{}
         \label{fig:acc_frames_actNet}
     \end{subfigure}
     \hfill
     \begin{subfigure}[b]{0.46\textwidth}
         \centering
         \includegraphics[width=\textwidth]{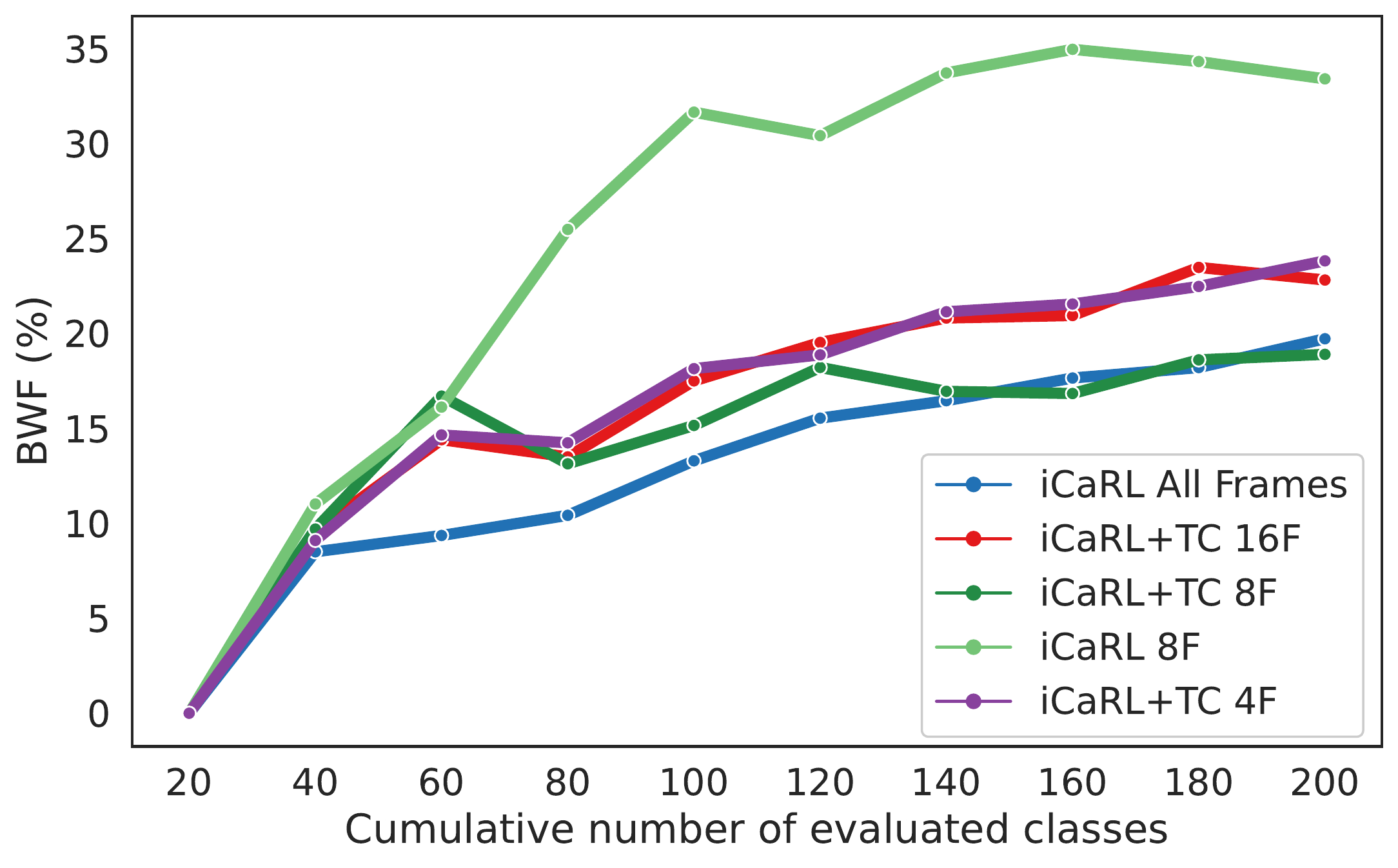}
         \caption{}
         \label{fig:bwf_frames_actNet}
     \end{subfigure}

\caption{
\textbf{Consistency Regularization Improves the Performance in the Validation Set of ActivityNet-Trim.} iCaRL experiences a sharp increase in backward forgetting when trained with memories of severely down-sampled videos (an example is shown here with 8 frames, but more examples are reported in Table 4 of the paper.) Our temporal consistency (TC) approach significantly alleviates this forgetting. Storing only 8 frames per video in memory, iCaRL trained with TC reports similar average accuracy and backward forgetting to the baseline iCaRL, which uses full-resolution videos for memory replay.
}
\label{fig:frames_actNet}
\end{figure}

\begin{figure}[t]
     \centering
     \begin{subfigure}[b]{0.46\textwidth}
         \centering
         \includegraphics[width=\textwidth]{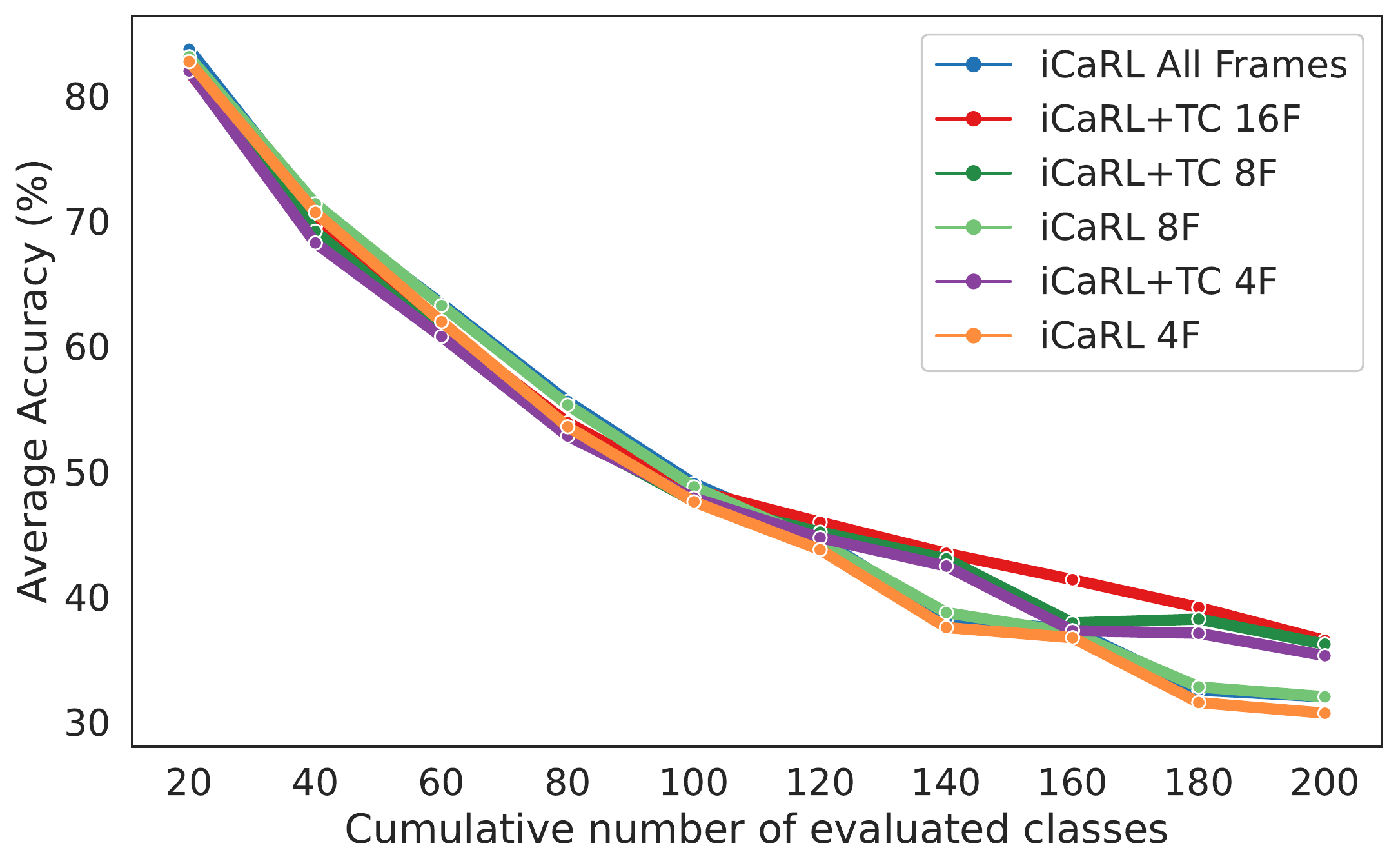}
         \caption{}
         \label{fig:acc_frames_kinetics}
     \end{subfigure}
     \hfill
     \begin{subfigure}[b]{0.46\textwidth}
         \centering
         \includegraphics[width=\textwidth]{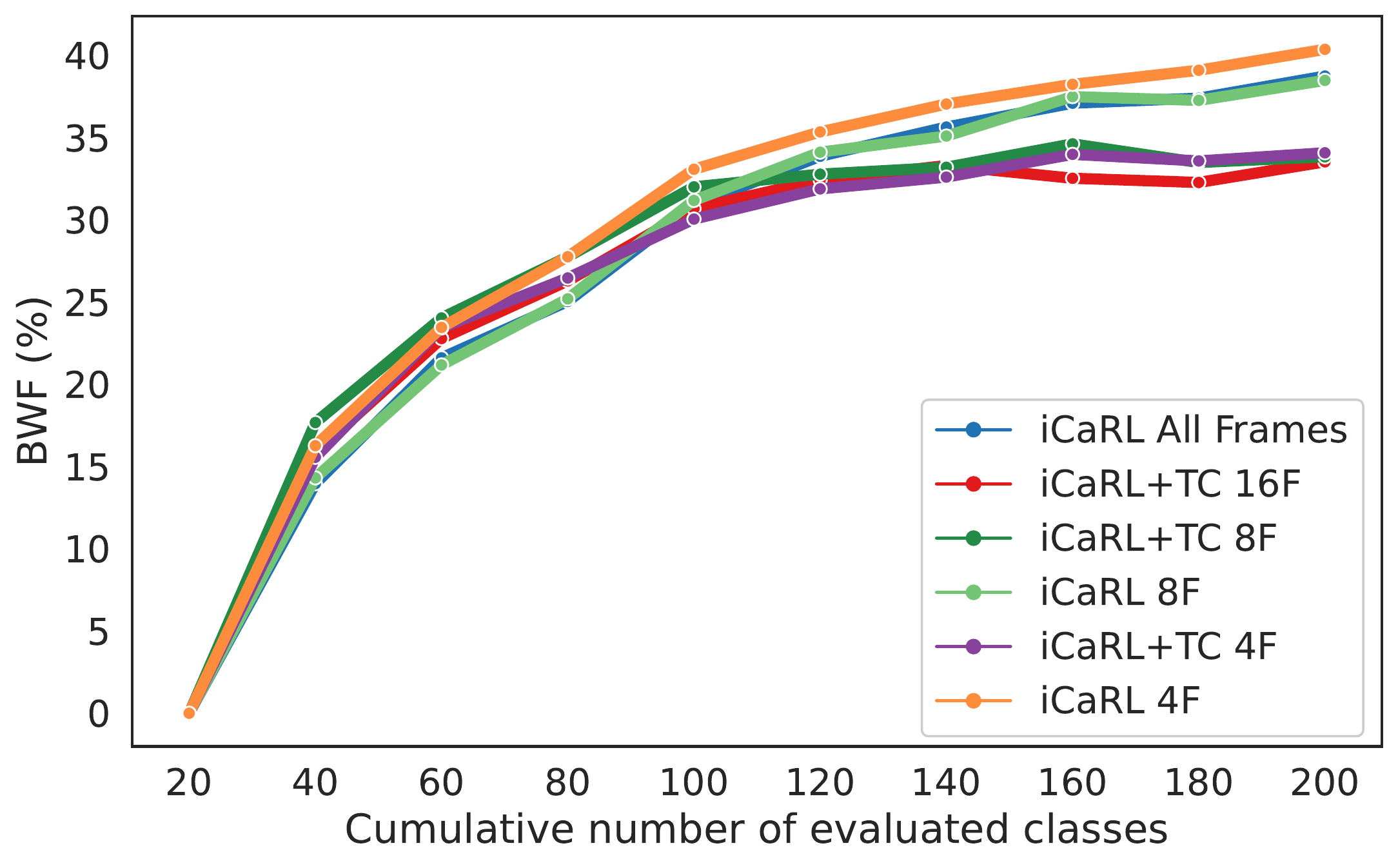}
         \caption{}
         \label{fig:bwf_frames_kinetics}
     \end{subfigure}

\caption{
\textbf{Consistency Regularization Improves the Performance in the Test Set of Kinetics.} iCaRL experiences the same increase in backward forgetting when trained with memories of severely down-sampled videos on Kinetics (an example is shown here with 8 frames, but more examples are reported in Table 3 of the paper.) Our temporal consistency (TC) approach significantly alleviates this forgetting. Storing only 8 frames per video in memory, iCaRL trained with TC reports even higher average accuracy than the baseline iCaRL, which uses full-resolution videos for memory replay.
}
\label{fig:frames_kinetics}
\end{figure}

As Figures \ref{fig:frames_actNet} and \ref{fig:frames_kinetics} show, in less-controlled scenarios like ActivityNet-Trim, which have long videos, the Naive and iCaRL methods are more susceptible to the number of stored frames per video than when they deal with short videos like in Kinetics. It is precisely in these scenarios where our TC loss is essential to reduce the memory consumption without affecting the model performance. In this scenario, we reduce approximately 99.79\% in memory usage leveraging the TC loss. 

\begin{table*}[t]
\centering
 \small
\resizebox{17cm}{!}{
\begin{tabular}{ccccccc} 
    
    \toprule
    \multicolumn{1}{c}{\multirow{2}{*}{\bf Model}} & \multicolumn{1}{c}{\multirow{2}{*}{ \begin{tabular}[c]{@{}c@{}}\textbf{Frames} \\ \textbf{per video} \end{tabular}}} & 
    \multicolumn{1}{c}{\multirow{2}{*}{ \begin{tabular}[c]{@{}c@{}}\textbf{Mem.} \\ \textbf{Frame Capacity} \end{tabular}}}
    & \multicolumn{2}{c}{\bf ActivityNet-Untrim} & 
    \multicolumn{2}{c}{\bf ActivityNet-Trim}
    \\\cmidrule(lr){4-7}
         & & & \multicolumn{1}{c}{\bf Acc $\uparrow$} & \multicolumn{1}{c}{\bf BWF $\downarrow$} & \multicolumn{1}{c}{\bf Acc $\uparrow$} & \multicolumn{1}{c}{\bf BWF $\downarrow$}\\
    \midrule
    Naive & 4 & \Scale[0.94]{1.6\times10^4} & 15.67\% & 32.97\% & 22.27\% & 35.24\%
    \\
    Naive & 8 & \Scale[0.94]{3.2\times10^4} & 18.25\% & 31.71\% & 22.44\% & 36.06\%
    \\
    \midrule
    Naive+TC & 4 & \Scale[0.94]{1.6\times10^4} & 37.58\% & 23.93\% & 43.11\% & 23.58\%
    \\
    Naive+TC & 8 & \Scale[0.94]{3.2\times10^4} & 39.80\% & 19.59\% & 43.97\% & 20.67\%
    \\
    \bottomrule
    \end{tabular}
}
\caption{
\textbf{Ablation study results with trimmed and untrimmed videos}. This table complements Table 4 of the main paper. All the experiments involve sequentially training on 10 tasks. ActivityNet-Untrim provides a more realistic and challenging setup to evaluate CIL models. We impose the strict resource constraint of 4 and 8 frames per video stored in memory. Results were shown in the paper for the iCaRL baseline. Here, we present the results for the Naive baseline, which selects memory samples randomly. Our temporal consistency approach improves the accuracy of both baseline methods trained with limited memory on ActivityNet-Trim and ActivityNet-Untrim by large margins. For Naive trained on 4 frames per memory video, TC results in a 22\% improvement. 
}
\vspace{-4mm}
\label{tab:naive_results}
\end{table*}

\subsection{What Actions Require Temporal Consistency (TC) to Learn in the Class Incremental Setup of ActivityNet-Trim?}
\begin{figure*}[t]

\centering
\includegraphics[width=1\linewidth]{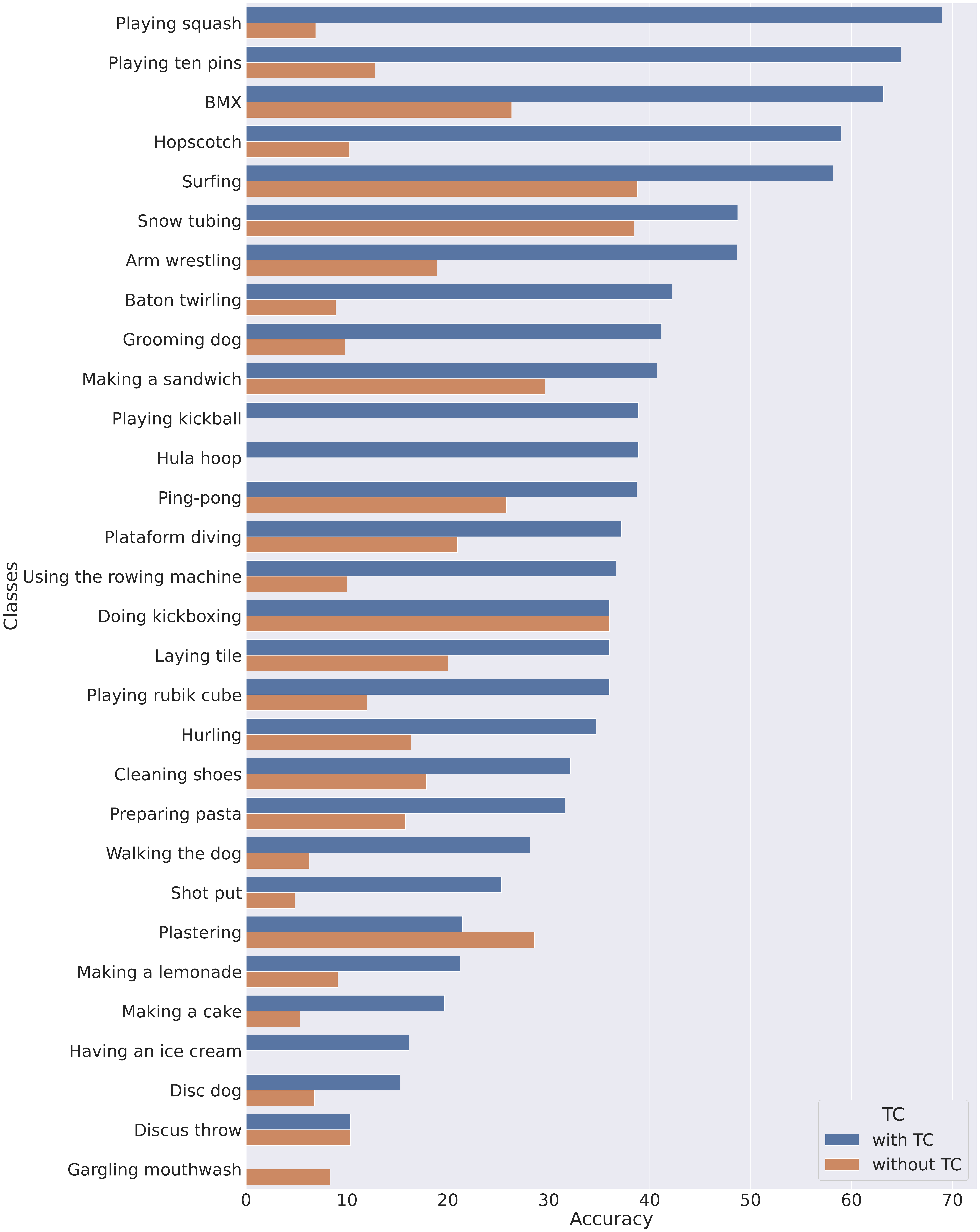}
\vspace{-3mm}
\caption{
\textbf{Temporal Consistency Training Improves iCaRL's Ability to Recognize Challenging Actions.} A more temporally consistent model is better able to recognize actions with fast movements, like playing squash and ten pins (bowling.)
}
\vspace{-6mm}
\label{fig:class_acc}
\end{figure*}

We analyze the per-class performance of iCaRL in the class incremental learning scenario, with the constraint of storing 8 frames per video in memory. We train two iCaRL models, one trained with TC and the other without TC, sequentially on 10 tasks of ActivityNet-Trim. Figure \ref{fig:class_acc} visualizes the performance of the two models on a sample of challenging classes, ordered by the performance of the TC model. Using our TC approach consistently and significantly improves the performance on most of the sampled classes. For some classes that require more temporal and motion reasoning, like playing kickball and hula hoop, the model without TC completely fails. Incorporating the TC loss improves the accuracy by large margins for these actions. Similarly, while iCaRL without TC achieves an accuracy of less than $8\%$ on the action of playing squash, training iCaRL with the TC loss results in an an accuracy of around $68\%$ on playing squash, enhancing the performance on that category by $60\%$.


